\pdfoutput=1

\documentclass[11pt]{article}

\usepackage
{acl}

\usepackage{newtxtext} 
\usepackage{latexsym}
\usepackage{graphicx}
\usepackage{booktabs}
\usepackage{amsmath}
\usepackage{amsfonts}
\usepackage{setspace} 
\usepackage{url}
\usepackage[vlined]{algorithm2e}
\usepackage{xspace}
\usepackage{xcolor}
\usepackage{multicol}

\usepackage[T1]{fontenc}
\usepackage[utf8]{inputenc} 

\usepackage{microtype}
\usepackage{wasysym}

\definecolor{Yellow}{HTML}{999900}
\definecolor{Lightblue}{HTML}{0000FF}
\definecolor{Blue}{HTML}{000099}
\definecolor{Red}{HTML}{990000}
\definecolor{Green}{HTML}{006600}

\newcommand\MLObj{\textit{MLE}\xspace}
\newcommand\tokenObj{\textit{Token\-Align}}
\newcommand\seqObj{\textit{Seq\-Align}}
\newcommand\seqRandomObj{\textit{SRand}}
\newcommand\seqCEObj{\textit{SeqAlign-CE}}
\newcommand\seqDecObj{\textit{SeqAlign-dec}}
\DeclareMathOperator*{\argmax}{arg\,max}
\newcommand{\BM}[1]{\ensuremath{\boldmath#1}}
\newcommand{\AlignmentA}{\textcolor{Blue}{Alignment~\textsf{\textit{A}}}\xspace}
\newcommand{\alignmentA}{
{alignment~\textsf{\textit{A}}}\xspace}
\newcommand{\AlignmentsA}{\textcolor{Blue}{Alignments~\textsf{\textit{A}}}\xspace}

\title{Soft Alignment Objectives for Robust Adaptation of Language Generation}
\author{Michal Štefánik$^{\clubsuit *}$ \and
  Marek Kadlčík$^\clubsuit$ \and
  Petr Sojka$^\clubsuit$ \\
  \\$^\clubsuit$Faculty of Informatics,\\\vspace{10pt}Masaryk University, Czech Republic
}
\begin{document}
\hyphenation{re-inforce}
\maketitle
\def\thefootnote{*}\footnotetext{Corresponding author: stefanik.m@mail.muni.cz}
\def\thefootnote{\arabic{footnote}}
\begin{abstract}
Domain adaptation allows generative language models to address specific flaws caused by the domain shift of their application.
However, the traditional adaptation by further training on in-domain data rapidly weakens the model's ability to generalize to other domains, making the open-ended deployments of the adapted models prone to errors.
This work introduces novel training objectives built upon a semantic similarity of the predicted tokens to the reference.

Our results show that
(1)~avoiding the common assumption of a single correct prediction by constructing the training target from tokens' semantic similarity can largely mitigate catastrophic forgetting of adaptation, while
(2)~preserving the adaptation in-domain quality,
(3)~with negligible additions to compute costs.
In the broader context, the objectives grounded in a continuous token similarity pioneer the exploration of the middle ground between the efficient but na\"{\i}ve exact-match token-level objectives and expressive but computationally- and resource-intensive sequential objectives.
\end{abstract}

\section{Introduction}

Large language models (LLMs) based on instances of encoder-decoder architecture \cite{Neyshabur2015InSO} provide a strong standard for generative applications of NLP, such as summarization or machine translation, mainly thanks to their outstanding ability to fluently model language.
These models might face issues with \textit{adequacy} of the generated text \cite{ustaszewski-2019-exploring} when applied in data domain(s) different from the training domain, but such errors can be partially mitigated using domain adaptation~\cite{Saunders2021DomainAA}.

Identically to the pre-training phase, the adaptation is commonly carried out using Maximum Likelihood Estimation (\MLObj) objective with teacher forcing~\cite{bahdanau2014neural}.
The popularity of this approach can be rightfully attributed to its outstanding data and computing efficiency.
However, model adaptation using \MLObj notoriously comes for a price of over-specialization to the target domain, also referred to as \textit{catastrophic forgetting}~\cite{Goodfellow2014AnEI}, characterized by a continuous decay of model performance on the inputs from the \textit{other} domains than the adaptation domain.

We hypothesize that catastrophic forgetting might be related to \MLObj{}'s na\"{\i}ve single-truth assumption, penalizing models' uncertainty over the possibly valid predictions, such as the synonyms. 
In domain adaptation, a repeated penalization of possibly valid tokens that are uncommon in the adapted domain might drive the model to unlearn the original features robust to meaning-invariant formulations.

\begin{figure}[t]
\!\!\!\centerline{\includegraphics[width=1.08\linewidth]{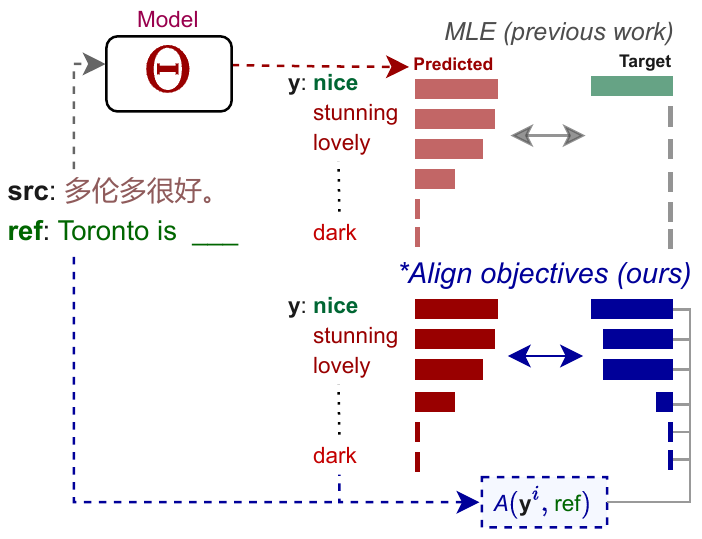}}%
\caption{\textbf{Soft alignment objectives} (\textcolor{Blue}{\textit{*Align}}) replace the single-truth assumption of Maximum Likelihood Estimation (MLE) objective by constructing target distribution using \AlignmentA 
based on the mutual similarity of token representations. We show that learning to model ambiguity in prediction can largely mitigate the loss of generalization in adaptation.}
\label{fig:abstract}
\end{figure}


We propose to counteract the single-truth assumption of \MLObj{} by constructing targets that respect mutual tokens' similarity through the alignment of output tokens to the reference (Figure~\ref{fig:abstract}). 
Consequentially, the expected target distribution is spread over the tokens that can be accurately aligned to the reference, based on the representations provided by a domain-agnostic embedding model.
We find that using such objectives in domain adaptation can eliminate a major portion of model performance loss on out-of-domain (OOD), caused by the adaptation techniques while reaching comparable or higher qualitative gains on the adapted domain.

Our main contributions are the following.
(i)~We present a framework for training generative language models with an alternative training signal based on token similarity provided by an arbitrary embedding model. A similar methodology can be applied for more robust training and adaptation of any language model.
(ii)~We introduce efficient and accurate training objectives that alleviate catastrophic forgetting of low-resource domain adaptation in NMT without losing adaptation quality.
(iii)~We further investigate the covariates that impact the robustness of generative LLM. Among others, we find that a more robust model can be obtained merely by exposing a generative model to its own predictions during the training.

This paper is structured as follows. Section~\ref{sec:background} surveys and compares our work to the existing work in training and adapting robust generative LLMs. Section~\ref{sec:objectives} introduces two main objectives that we experiment with: \tokenObj{} and \seqObj{}. Section~\ref{sec:methodology} describes our experimental methodology and ablation analyses and Section~\ref{sec:results} summarizes our findings, highlighting the broader implications.






\section{Background}
\label{sec:background}

Language generation is the modus operandi for a wide set of problems requiring an open-ended sequence of tokens as the answer. Machine translation is the representative of this group that we focus on, but other tasks such as summarization \cite{lewis2020bart}, vision captioning \cite{arxiv.2203.15350}, question answering \citep{t5} or in-context learning \cite{Sanh2021LearningFO} are also applications of the described framework.

In the commonly-used auto-regressive generation, for each pair of input and reference sequence of tokens $X_j$ and $Y_j$, a \textit{language model} $\Theta(Y_{j, i} | X_j, Y_{j, 1..i-1})$ is trained to generate output sequence by maximizing the probability of generating the $i$-th token $y_{ji} = \argmax(\Theta(X_j, Y_{j, 1..i-1}))$ \textit{matching} the reference $Y_{ji}$ while minimizing the probability of generating other tokens of the vocabulary, conditionally to the input text $X_j$ and \textit{previous} reference tokens~$Y_{j,1..i-1}$:
\begin{equation}
    \max p(y_{ji} = Y_{ji}|X_j, Y_{j,1..i-1}, \Theta)
\end{equation}

This objective is implemented in the commonly-used Maximum Likelihood Estimation (\MLObj{}) objective, which minimizes a cross-entropy (CE) of the predicted distribution of $\Theta(X_j, Y_{j,1..i-1})$ to the \textit{expected} distribution, which is a one-hot encoding $E_{ji}$ of the \textit{true} reference token $Y_{ji}$ over the model vocabulary:
\begin{equation}
    \mathcal{L}_{\textit{MLE}}(\Theta) = \min\left(\! - \log \frac{\exp(\Theta(X_j, Y_{j,1..i-1}))}{\exp(E_{ji})}\!\right)
    \label{eq:ml}
\end{equation}

This objective is commonly used both for training \citep{cho,attention} and adaptation \citep{Servan2016DomainSA,Saunders2021DomainAA} of generative LLMs.



While the adaptation brings benefits in modelling domain-specific terminology \cite{sato-etal-2020-vocabulary}, or in avoiding inadequate generation artefacts such as repetitions or hallucinations \cite{etchegoyhen-etal-2018-evaluating}, it comes at a price of generalization to other domains;
the adapted models improve on the adapted domain but gradually perform worse on other domains.

Previous work in domain adaptation presents methods addressing the mitigation of catastrophic forgetting.
\citet{chu-etal-2017-empirical} enhance model robustness by mixing the pre-training and adaptation samples in continuous training, assuming that the full pre-training dataset is available, which is commonly not the case.
\citet{thompson-etal-2019-overcoming} regularize the training objective with Fischer Information Matrix. 
\citet{dakwale2017finetuning} also use the regularization in training, instead based on the predictions of the original model.
Similarly, \citet{Freitag2016FastDA} use the ensemble of the original and trained model in prediction. In this line, we experiment with the ensemble approach using Transformers but find it underperforms other methods in low-resource adaptation.

\citet{han-etal-2021-robust} find that using parameter-efficient fine-tuning methods, such as using \textit{Adapters} \citep{pmlr-v97-houlsby19a} can increase the robustness of the adapted model. 
Previous work also applied Adapters in the fine-tuning of generative LLMs \citep{cooper-stickland-etal-2021-multilingual,lai-etal-2022-m4Adapter}, but do not evaluate the distributional robustness of the final models; 
Therefore, we include Adapters as another baseline, but find it also struggling in lower-resource cases, due to the random initialisation of its bottleneck representations. We find that this problem can be avoided using \textit{LoRA} \cite{hu2022lora}, which instantiates tuned parameters as \textit{additions} to attention matrices initialised close to zero values, therefore commencing the adaptation with the originally-performing model.



Another problem of \MLObj{} is referred to as \textit{exposure bias}: while in the teacher-forced training, the model's $i$-th prediction $\Theta(X_j)_i$ is conditioned by the correctly-generated previous tokens from the reference $Y_{j,1..i-1}$, in practice, the model conditions its predictions on its \textit{own} outputs $\Theta(X_j)_{1..i-1}$.
We speculate that this discrepancy might be magnified under a domain shift where the model can not learn to follow references in a generation.

Exposure bias was addressed by \textit{sequential objectives}, such as Minimum Risk Training (MRT) \citep{Ranzato2016SequenceLT} that optimize the model by the evaluation of complete output sequence \cite{yang-etal-2018-improving,wang-sennrich-2020-exposure,9099374,JauregiUnanue2021BERTTuneFN}.
Apart from the specifics of Reinforcement learning, such as fragility to the optimization settings \cite{JMLR:v22:20-303}, these methods are also more resource-demanding as they require a sequence of predictions for a single update, limiting their applicability in low-resource adaptation.
Previous work of \citet{DBLP:conf/iclr/ChoshenFAA20} also shows that gains of sequential methods in adaptation might be similar to a random training signal.
Inspired by this finding, we also assess the gains and OOD robustness of our methods against a random-feedback sequential baseline~(§\ref{sec:ablation}).\looseness=-1

Closer to us, previous work uses alternative training signal based on comparing model hypotheses to the reference.
\citet{xu-etal-2019-differentiable} build soft alignment between fully-generated hypotheses based on hidden states of bidirectional LSTM encoder-decoder and weigh the predicted probability distribution by such alignment in the training objective.
Similarly, \citet{lu-etal-2020-mixed} complement \MLObj{} and sentence-level objective with the objective minimizing a dot-product of the best-matching hidden representations of tokens of a hypothesis and a reference. 
\citet{Chen2019ImprovingSL} and later \citet{Zhang2020SemanticMF} introduce the matching scheme that uses the Optimal transport cost \citep{kusner2015from} of the embeddings of reference to the hypothesis as their objective loss.

Referenced work reaches improvements in conventional high-resource training scenarios, whereas our goal is to propose a method for training robust generative models for challenging low-resource settings.
This also motivates a primary difference in the design of our methods; That is, to use domain-agnostic representations for constructing training targets, instead of the model's own representations, which are subject of over-specialization in adaptation.






\section{Soft Alignment Objectives}
\label{sec:objectives}

This section describes the details of alignment-based objectives\footnote{The implementation of all new objectives is available at:\\ \url{https://github.com/MIR-MU/softalign_objectives}} that we introduce in this work.

\subsection{Token Alignment}
\label{sec:alignment}

\begin{figure}
  \centerline{\includegraphics[width=1.0\linewidth]{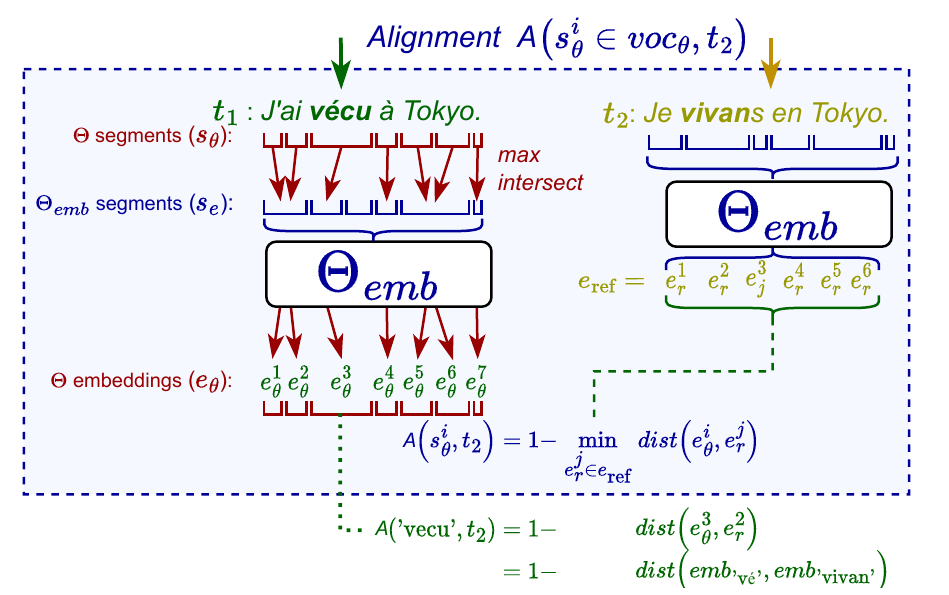}}
    \vspace*{-.0\baselineskip}
  \caption{\textbf{Token alignment mechanism} represents subwords $s_\Theta$ of the trained model $\Theta$ with embeddings of a robust, static model \textcolor{Blue}{$\Theta_\textit{emb}$}. 
  Using these representations, we define \textit{Alignment} of any $\Theta$'s subword \textcolor{Green}{$s_\Theta^i$} to another text \textcolor{Yellow}{$t_2$} through a minimal distance of their embeddings given by the robust embedding model \textcolor{Blue}{$\Theta_\textit{emb}$}.}
  \label{fig:alignment_scheme}
\end{figure}

Our goal is to circumvent the single-truth assumption of \MLObj{} with targets respecting the mutual tokens' similarity. Since the representations of the trained models are affected by catastrophic forgetting, we propose to use an alternative, domain-agnostic representation model ($\Theta_\textit{emb}$) to provide the token representations, i.e.\ embeddings.

However, as the vocabularies of the fine-tuned model $\Theta$ and $\Theta_\textit{emb}$ are not aligned, to train with representations of a different $\Theta_\textit{emb}$, we need to \textit{match} each subword (token) of the trained model ($s_\Theta^i$) with a subword of the embedding model ($s_e^j$) having a representation $e^j \in \Theta_\textit{emb}(t)$;
(i)~We tokenize input text $t_1$ using both $\Theta$'s and $\Theta_\textit{emb}$'s tokenizers, obtaining subwords $s_\Theta$ and $s_e$ respectively.
(ii)~Then, we compute the character-level positional spans of both subwords lists $s_\Theta$ and $s_e$. Finally, we (iii) \textit{match} each model subword $s_\Theta^i \in s_\Theta$ with embedding subword $s_e^j \in \Theta_\textit{emb}$ such that $s_e^j$ has the largest positional overlap with $s_\Theta^i$.
As a result, each $\Theta$'s subword $s_\Theta^i$ gets assigned an embedding $e_\Theta^i = e_r^k$ of $\Theta_\textit{emb}$, as visualized in Figure~\ref{fig:alignment_scheme}.

Having $\Theta$'s subwords' representations from a robust embedding model $\Theta_\textit{emb}$, we finally define an \textit{Alignment} $\mathcal{A}$ of any subword $s_\Theta^i \in \Theta$ to another text $t_2$ as:
\begin{equation}
    \mathcal{A}(s_\Theta^i, t_2) = 1 - \min_{e_r^j\in \Theta_\textit{emb}(t_2)} \textit{dist}(e_\Theta^i, e_r^j)
    \label{eq:align}
\end{equation}
where $\textit{dist}$ is any distance measure defined for the chosen embedding system.
In our experiments, we use standard Euclidean distance as the measure.
We provide a more detailed description and complexity analysis of the Alignment algorithm $\mathcal{A}$ in Appendix~\ref{appx:alignment}.

\subsection{\tokenObj{} Objective}
\begin{figure}
  \centerline{\includegraphics[width=0.95\linewidth]{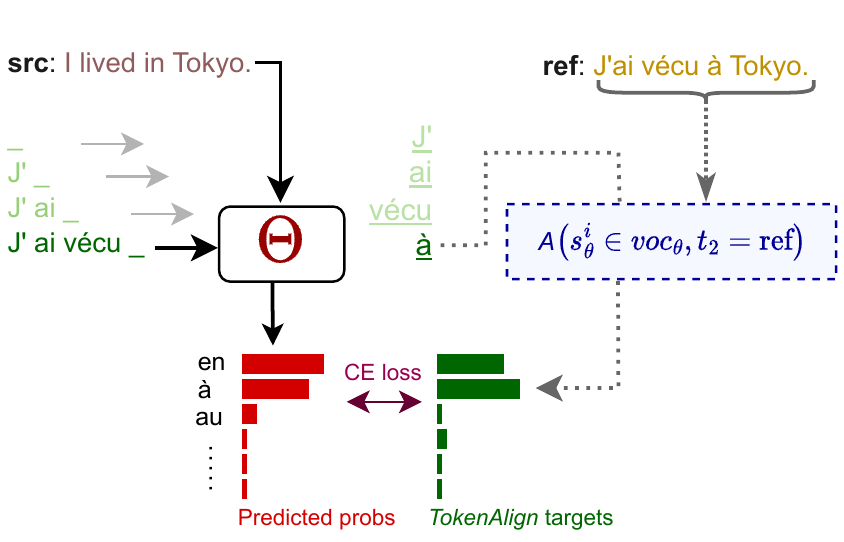}~~~~~}
  \vspace*{-.3\baselineskip}
  \caption{\textbf{\tokenObj{} objective} replaces one-hot targets of \MLObj{} with token \AlignmentsA based on a similarity between the embeddings of the candidate and reference tokens (§\ref{sec:alignment}), encouraging the trained model \textcolor{Red}{$\Theta$} to respect the ambiguity of prediction, instead of eliminating it.}
  \label{fig:tokenalign}
\end{figure}

\label{sec:tokenobj}

\tokenObj{} is designed as a minimal adjustment to \MLObj{} (Eq.\ \eqref{eq:ml}) using the \alignmentA as the target of each candidate token of $\Theta$'s vocabulary. Instead of penalisation, this encourages the model to up-weight predictions that do not match the reference token, but still can be accurately matched to the reference text (Figure~\ref{fig:tokenalign}):
\begin{equation}
    \!\!\!\!\mathcal{L}_{\textit{TAlign}}(\Theta) = \min\!\left(\!\! - \log
    \frac{\exp(\Theta(X_j, Y_{j,1..i-1}))}
    {\exp(\mathcal{A}(\textit{voc}_\Theta, Y_j))}\!\right)
    \label{eq:talign}
\end{equation}
where $\textit{voc}_\Theta$ is the vocabulary of $\Theta$, and $\mathcal{A}(s_\Theta^{1..|\Theta|}, Y_j)$ are the \textit{alignments} for each token of the vocabulary ($s_\Theta^i$) to the reference text $Y_j$. Note that none of $\mathcal{A}$'s components is updated in training.

Relying on the same training approach as with the conventional \MLObj{} objective, \tokenObj{} presents an alternative of the \MLObj{} of similar data and compute efficiency (compared in Appendix~\ref{appx:speed}). However, \tokenObj{} still does not address the exposure bias as the model $\Theta$ is only updated conditionally to the previous \textit{reference} tokens $Y_{1..i-1}$ as the prefixes, rather than its own outputs.

\subsection{\seqObj{} Objective}
\label{sec:seqobj}
\begin{figure}
  \centerline{\includegraphics[width=1.02\linewidth]{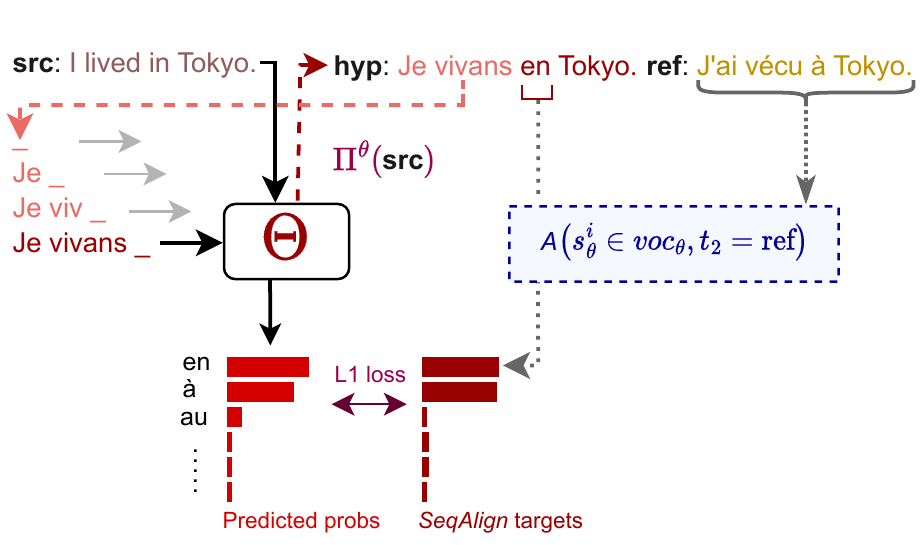}}
  \vspace*{-.3\baselineskip}
  \caption{\textbf{\seqObj{} objective} further replaces the reference prefixes in the training with \color{Red}$\Theta$\color{black}'s own-generated hypotheses. This additionally adapts the model to condition the predictions based on its own outputs, instead of the reference.}
  \label{fig:seqalign}
\end{figure}

Alignment $\mathcal{A}$ allows us to assess $\Theta$'s prediction quality on a token level, but without dependence on the exact ordering of reference tokens. Thus, we no longer need to keep the prefixes synchronized with reference and can construct targets for an arbitrary prefix.
Hence, instead of taking prediction prefixes from reference $Y_j$, \seqObj{} constructs the prefixes from the hypothesis generated by the trained model~$\Theta$ itself (Fig. \ref{fig:seqalign}).



We create the self-generated \textit{hypothesis} by using $\Theta$'s outputs as a probability distribution and construct a generation strategy $\Pi^\Theta$ that \textit{samples} next token(s) from this distribution.
A desideratum of such generation strategy (compared to a greedy search) is that the prefixes of generated hypotheses are diverse but still realistically likely to occur during $\Theta$'s generation.

Additionally, instead of generating a \textit{single} hypothesis for each input, we can obtain a \textit{set of hypotheses} $\hat{Y}_j \sim \Pi^\Theta(X_j)$ that can be used by \seqObj{} to condition the updates of $\Theta$.
The sampling generation strategy is inspired by the previous work, using sampling to construct full hypotheses \citep{neubig-2016-lexicons,Shen2017OptimizingNE,edunov-etal-2018-classical}.

Identically to \tokenObj{}, \seqObj{} associates \textit{all} the vocabulary tokens $\textit{voc}_\Theta$ with their alignment quality $\mathcal{A}(s_\Theta^{1..|\Theta|}, Y_j)$ and uses the alignment as target distribution.
However, motivated by the empirical results, instead of the Cross-Entropy, we minimise \textit{absolute distance} ($L1$) as \seqObj{}'s training objective:
\begin{equation}
    \begin{aligned}
        \!\mathcal{L}_{\textit{SAlign}}(\Theta) &\!=\! \min \!\left(\Theta(X_j, \hat{Y}_{j,1..i-1}) - \mathcal{A}(\textit{voc}_\Theta, \!Y_j)\right) \\
        \text{where } \hat{Y}_j &\sim \Pi^\Theta(X_j)\\[-3mm]
    \end{aligned}
    \label{eq:salign_dist}
\end{equation}

Note that we further analyse the impact of the loss formulation in the ablation in Section~\ref{sec:ablation:loss_formulation}.

\subsection{Embeddings Contextualization}
\label{sec:decontextualization_intro}

Computing alignment $\mathcal{A}$ using \textit{context-insensitive} embedding model $\Theta_\textit{emb}$, such as GloVe \cite{pennington-etal-2014-glove} or FastText \citep{bojanowski2017enriching} requires no further adjustments. However, using more expressive \textit{context-sensitive} embedding models, such as BERT \citep{devlin2018bert} for computing $\mathcal{A}$ as a target for any possible output token faces the following issues.

(i)~Inference of representations \textit{on the fly} within the training process is expensive. Consider an example of obtaining contextual representations for each possible next token in generating a 10-token hypothesis, requiring $10^{|\Theta|}$ inferences of $\Theta_\textit{emb}$, where $|\Theta|$ is a size of the vocabulary of $\Theta$, commonly in ranges of 30,000--50,000 tokens.

(ii)~A full context required to infer bidirectional contextual embeddings remains incomplete throughout the generation. The embeddings could be inferred within a synthetic context or
using a unidirectional embedding model instead, but we find that both these approaches significantly alter tokens' pairwise distances.

In the \seqObj{} objective, we address these issues by embedding only the top-$n$ \textit{highest-scored tokens} of $\Theta$ in each prediction step (denoted $\Theta^{\uparrow n}$).
By fixing $n=3$, we need to infer the contextual embeddings of only $\sum_{k=1}^K 3|\Pi_k(X_j)|$ of the highest-scored tokens for each sampled hypothesis $\Pi_k(X_j)$.
In our experiments, we also keep the \textit{number of sampled hypotheses} $K$ fixed to $K=10$
and we do \textit{not} adjust $\Theta$ by the scores of the tokens other than the top ones.
As the context, we use the complete hypothesis from which the token $s_\Theta^i \in \Theta^{\uparrow n}$ is sampled.
Therefore, the targets $\mathcal{A}$ for our distance-based objectives are adjusted to:
\begin{equation}
    \mathcal{A'}(s_\Theta^i, t_2) =
    \begin{cases}
        \mathcal{A}(s_\Theta^i, t_2)& \text{if } s_\Theta^i \in \Theta^{\uparrow n}\\
        0              & \text{otherwise}
    \end{cases}
    \label{eq:contextual_align}
\end{equation}

In \tokenObj{}, which requires embeddings of \textit{all} tokens of the vocabulary, we address the computational overhead in a \textit{decontextualization} process.
We obtain the decontextualized embedding $e^i$ for each subword $s_e^i$ as an \textit{average} of the contextualized embeddings corresponding to \textit{all} the occurrences of $s_e^i$ in the texts of the training domain~$X$:
\begin{equation}
    e_\textit{dec}^i = \frac{1}{\text{\#}s_e^i} \sum_{\substack{X_j \in X;\, s_e^i \in X_j}} \!\!\!\!\!\!\!\!\!\Theta_\textit{emb}(X_j)^i
    \label{eq:dec_embeddings}
\end{equation}
where \#$s_e^i$ is the number of occurrences of a subword $s_e^i$ in~$X$.

While such a process also causes qualitative decay of the contextual representations, it has been shown that decontextualized representations still outperform context-agnostic (FastText) embeddings in machine translation evaluation \cite{stefanik-etal-2021-regressive}.
Despite that, we quantify the impact of decontextualization as one of our ablations (§\ref{sec:ablation}).

Throughout all our experiments, we use the embeddings of multilingual BERT model \cite{devlin-etal-2019-bert} as $\Theta_\textit{emb}$, extracted from the 9-th hidden layer, motivated by the previous work of \citet{bertscore} showing this model to best correlate with a human evaluation of generative LLMs.

\section{Experiments}
\label{sec:methodology}

We evaluate the impact of the proposed training objectives in the domain adaptation experiments in machine translation, where the distributional robustness in adaptation may bring well-measurable benefits. We compare our results with the adaptation using the commonly-used \MLObj{} objective (§\ref{sec:background}), and selected parameter-efficient methods shown to mitigate forgetting.
We use the novel objectives as the weighted \textit{complements} of the \MLObj{} objective (Eq.\ \eqref{eq:ml}), aiming to extend the modelled space of the problem complexity:
\begin{equation}
     \mathcal{L}_{\textit{*Align}}(\Theta) = \mathcal{L}_{\textit{MLE}}(\Theta) + \alpha \cdot \mathcal{L}_{\textit{NewObj}}(\Theta)
     \label{eq:weighting}
\end{equation}

\subsection{Datasets}
\label{sec:datasets}

We choose the data configurations of our experiments to allow the reader to extrapolate trends and conclusions invariant to the following covariates.

\textbf{Domains.} To assess the distributional robustness of the models, we train and evaluate among \textit{all} pairs of these OPUS domains \citep{tiedemann-2012-parallel}: \textit{Wikimedia}, \textit{OpenSubtitles}, \textit{Bible}, \textit{TEDTalks}, \textit{DGT/Law} and \textit{EMEA/Medical}.
We choose the set of domains that reflects both minor (\textit{Wikimedia} → \textit{OpenSubtitles}) and major (\textit{EMEA/Medical} → \textit{Bible}) domain shifts between the training and evaluation. Our selection reflects on real-world settings where practitioners commonly adapt a general-purpose model to a \textit{specialized} domain such as \textit{law} or \textit{medicine}, but need to keep an operational level of quality on any input.

\textbf{Data size.} We focus on the applications where the size of parallel corpora available for adaptation range from \textit{very low-resource} (50,000 aligned sentences, \textit{Bible}) to \textit{medium-resource} (5,100,000 sentences, \textit{DGT/Law}).

\textbf{Language pairs.} Our evaluated language pairs are: \textit{Estonian → English}, \textit{German → English} \textit{English → Czech}, \textit{English → Ukrainian}, \textit{English → German} and \textit{English → Chinese}.
We pick the \textit{English-centric} pairs in order to maximize the number of out-of-domain evaluation sources for the adapted language pair.
Our settings cover target languages of Latin, Cyrillic, and Chinese alphabets.

\subsection{Experimental Setup}
\label{sec:experimental_setup}

\paragraph{Data configuration} As the OPUS sources do not contain standard splits, we split the data into train-validation-test. We first de-duplicate the samples and draw 500 validation and 1,000 test samples from each domain.

\paragraph{Training}
We perform the adaptations from the bilingual Transformer-base models of \citet{attention} using the checkpoints of \citet{tiedemann-thottingal-2020-opus} pre-trained for a translation of the corresponding language pair on a mixture of OPUS sources.

We perform a hyperparameter search over the parameters of \textit{learning rate}, \textit{objectives weights} $\alpha$, and objective-specific \textit{batch size}.
We detail the values and ranges of this search in Appendix~\ref{appx:hyperparameters}.

After fixing the objectives' parameters, we set up the experiments to closely resemble the traditional training process; We run each experiment until early-stopping by in-domain validation BLEU, with the patience of 20~evaluations, i.e., 10,000 updates and evaluate the model with the best validation score for testing. If the model does not improve over the first 10,000 updates, we evaluate the resulting model after the 10,000 updates.

We implement our experiments using Adaptor library \citep{stefanik-etal-2022-adaptor}, allowing the release of all our experiments in a transparent and self-containing form.\footnote{All our experiments can be reproduced by running a single line of code; refer to the Section \url{experiments} in \url{https://github.com/MIR-MU/softalign_objectives}}

\paragraph{Evaluation}
To discourage the effect of the random variance in the performance of the trained model, we report all test scores as the \textit{average} of the performance in the interval of 5~\textit{preceding} and 5~\textit{succeeding} checkpoints, resulting in a single, average test evaluation for each domain.

We collect evaluations of BLEU in the default settings of SacreBLEU \cite{post-2018-call}, obtaining a single (average) evaluation of in-domain (ID) BLEU and a set of corresponding evaluations for \textit{all} listed domains \textit{other} than the in-domain (OOD).
Given the availability of the sources, this results in four OOD evaluations for all pairs except (en→ukr) and (en→zh) with the datasets for two OOD evaluations.\looseness=-1

To enable mutual comparability, we finally normalize both ID and OOD results by the performance of the initial checkpoint and report the change of performance in percentage.
We report a single scalar value, or an interval in a form <\textit{mean}$\pm$\textit{range covering all results}>.

\paragraph{Baselines}
In addition to \MLObj{}, we compare the proposed methods to four existing methods reported to enhance LLMs'robustness.
(i)~Label smoothing \cite{Szegedy2016RethinkingTI} with $\alpha=0.1$ used widely also for training MT models distributes a constant portion of expected probability among all possible predictions.
(ii)~\textit{Adapters} \cite{pmlr-v97-houlsby19a} freezes pre-trained model parameters and fine-tunes a small set of newly-initialized bottleneck parameters.
Instead, (iii)~\textit{LoRA} avoids Adapters' issue of breaking the model in the initial training phase by initializing the new parameters that are trained as an \textit{addition} to the model's original, frozen parameters. (iv) We also implement and evaluate the Ensemble approach of \cite{Freitag2016FastDA}, but find this approach unable to bring adaptation gains in either of our relatively low-resource adaptation cases.
We detail the settings of our baselines in Appendix~\ref{appx:hyperparameters}.

\begin{table*}[tbh]
\centering\tabcolsep 4dd
\scalebox{0.68}{%
\begin{tabular}{@{}c@{\hspace{-3mm}}rllllllll@{}}
\toprule
  \multicolumn{1}{l}{} &
  \multicolumn{1}{l}{$\!\!\Delta$ BLEU} &
  \multicolumn{1}{l}{\begin{tabular}[c]{@{}l@{}} Bible\\ (de→en)\end{tabular}} &
  \multicolumn{1}{l}{\begin{tabular}[c]{@{}l@{}} TEDTalks\\ (en→zh)\end{tabular}} &
  \multicolumn{1}{l}{\begin{tabular}[c]{@{}l@{}} Opensubs\\ (en→ukr)\end{tabular}} &
  \multicolumn{1}{l}{\begin{tabular}[c]{@{}l@{}} Wiki\\ (en→cze)\end{tabular}} &
  \multicolumn{1}{l}{\begin{tabular}[c]{@{}l@{}} Medical/EMEA\\ (est→en)\end{tabular}} &
  \multicolumn{1}{l}{\begin{tabular}[c]{@{}l@{}} Law/DGT\\ (en→de)\end{tabular}} &
  \multicolumn{1}{l}{\begin{tabular}[c]{@{}c@{}} \textbf{Average} \\ \ (BLEU) \, \end{tabular}} &
  \multicolumn{1}{l}{\begin{tabular}[c]{@{}c@{}} \!\!\!\!\textbf{Average} \\ \!\!(BERTScr) \end{tabular}}
\\
  & & \small{62,000 pairs} & \small{155,000 pairs} & \small{877,000 pairs} & \small{1,003,000 pairs} & \small{1,021,000 pairs} & \small{5,105,000 pairs} & 
\\ \hline
  Orig. BLEU & & $21.89$ & $29.01$ & $26.12$ & $34.04$ & $54.85$ & $33.56$
\\ \midrule
  \MLObj{}    & ID        & \BM{-\,\ 8\%}         & \BM{+\,\ 7\%}                  & $+\ \, 4\%$                   & $+\,\ 9\%$                   & $+38\%$                  & $-\,\ 1\%$ & $+\,\ \textbf{8.31\%}$ & $+\,\ 9.19\permil$   
\\
 \cite{bahdanau2014neural}           & OOD       & $-53\%\pm36\%$           & $-23\%\pm23\%$            & $-15\%\pm9\%$            & $-15\%\pm5\%$            & $-35\%\pm10\%$           & $-19\%\pm11\%$        & $-26.87\%$  & $-37.34\permil$
\\
\addlinespace
  \MLObj{} + \textit{Smoothing}    & ID        & $-\,\ 6\%$         & $+\textbf{30}\%$                  & $-\ \, 6\%$                   & $+\,\ 9\%$                   & $+17\%$                  & $+\,\ 0\%$ & $+\,\ 7.43\%$ & $+\ \,3.77\permil$  
\\
 \cite{Szegedy2016RethinkingTI}  & OOD       & $-85\%\pm31\%$           & $-39\%\pm26\%$            & $-25\%\pm9\%$            & $-13\%\pm22\%$            & $-49\%\pm16\%$           & $-27\%\pm26\%$ &  $-41.86\%$ & $-54.13\permil$
\\ \addlinespace
  \textit{Adapters}    & ID        & $-\,\ \textbf{5}\%$         & \BM{-27\%}                  & $-14\%$                   & $+\,\ 1\%$                   & $+13\%$                  & $-\,\ 0\%$ & $-\,\ 5.41\%$ & $-15.23\permil$   
\\
 \cite{pmlr-v97-houlsby19a} & OOD       & $-91\%\pm20\%$           & $-80\%\pm2\%$            & $-53\%\pm9\%$            & $-46\%\pm25\%$            & $-77\%\pm19\%$           & $-45\%\pm43\%$  & $-65.39\%$   & $-94.97\permil$ 
\\  \addlinespace
  \textit{LoRA\ }    & ID        & \BM{-\,\ 8\%}         & \BM{+\,\ 2\%}                  & $+\ \, 2\%$                   & $+\textbf{14\%}$                   & $+\ \,8\%$                  & $+\,\ 6\%$ & $+\,\ 3.98\%$  & $+\ \,5.85\permil$   \\
 \cite{hu2022lora} & OOD       & $-\ \,7\%\pm7\%$           & $-21\%\pm20\%$            & $-\ \,\textbf{1\%}\pm1\%$            & $\ \,-7\%\pm5\%$            & $-\ \,4\%\pm11\%$           & $\,\ +2\%\pm14\%$ & $-\,\ 5.15\%$ & $-\ \,3.78\permil$     
\\ \midrule
  \textbf{\tokenObj{}} & ID        & $-21\%$                  & $+\ \,2\%$                   & $+\,\ \textbf{8\%}$          & \BM{+12\%}          & $+\textbf{45\%}$         & $+\,\ 1\%$ & $+\,\ 8.17\%$ & $+\ \,6.83\permil$       \\
  (ours)          & OOD       & $-\,\ 2\%\pm1\%$           & $-\textbf{10\%}\pm\!12$\%            & $-\,\ \textbf{1\%}\pm\!1\%$   & $-\,\ \textbf{6\%}\pm\!6\%$   & $-\,\ 6\%\,\pm\,\ 7\%$           & $+\ \,\textbf{6\%}\pm\!20\%$ & $-\,\ 3.25\%$ & $-\ \,\textbf{0.98}\permil$
\\ \addlinespace
  \textbf{\seqObj{}}   & ID        & $-23\%$                  & \BM{+\,\ 7\%}                   & $-\,\ 8\%$                   & $+\,\ 8\%$                   & $+31\%$                  & $+\ \,\textbf{7\%}$ & $+\,\ 3.67\%$ & $+\textbf{15.46\permil}$
  \\
  (ours)          & OOD       & $-\,\ \textbf{1}\%\pm\!1\%$ & $-20\%\pm22\%$            & $-\,\ 2\%\pm3\%$            & $-12\%\pm5\%$            & $-\,\ \textbf{1\%}\pm2\%$  & $+\,\ 3\%\pm13\%$ & $-\,\ \textbf{1.44\%}$ & $-\,\ 1.53\permil$
  \\
\bottomrule
\end{tabular}
}
\caption{\textbf{Evaluation of adaptation quality and robustness}: 
A~change of BLEU score relative to the original model, when adapting pre-trained Transformer on the titled domain, as measured on a held-out set of the training domain (in-domain, ID) and other listed domains available for the same language pair (out-of-domain, OOD). \textbf{Bold} denotes the best Average ID and OOD results, and per-domain results, where adaptation brings ID improvements. The results are evaluated using SacreBLEU \cite{post-2018-call} and BERTScore~\cite{bertscore}.}
\label{table:results_overview}
\end{table*}

\subsection{Ablation Experiments}
\label{sec:ablation}

In a set of additional experiments, we estimate the impact of the crucial components of the soft alignment objectives on adaptation accuracy and robustness. While these assessments provide an ablation study verifying our design decisions, they also assess the impact of different design aspects on the robustness of generative language models.

\paragraph{Impact of teacher forcing}

Teacher forcing, i.e.\ replacing the model's own outputs with the preceding tokens of the reference (§\ref{sec:background}) circumvents the problem of aligning the model's generated output to the reference.
We suspect that the discrepancy between the training and generation can be magnified under the distribution shift and hence, can be one of the causes of the catastrophic forgetting.

To assess the impact of teacher forcing on robustness, we design an objective that uses the model's generated outputs as prefixes, but contrary to \seqObj{}, it provides \textit{non-informative} training signal. We implement the experiment by replacing the \seqObj{}'s alignment $\mathcal{A}$ (in Eq.~\eqref{eq:salign_dist}) with \textit{randomly-generated} alignment $A_\textit{rand}$ as target:
\begin{equation}
   \mathcal{L}_{\textit{SRand}}(\Theta) = \min \!\left[\Theta(X_j, \hat{Y}_{j,1..i-1}) - \mathcal{A}_\textit{rand}\right]
    \label{eq:srandom}
\end{equation}

Additionally to the assessment of the impact of teacher forcing removal, this experiment also quantifies the importance of the embedding-based training signal of \seqObj{}.

\paragraph{Impact of decontextualization}
\label{sec:decontextualization}

While the \tokenObj{} utilize the \textit{decontextualized} grounding embeddings (§\ref{sec:decontextualization}), the decontextualization likely affects the quality of target distribution.
However, as we discussed in Section~\ref{sec:decontextualization_intro}, it is not computationally feasible to simply infer the contextualized embeddings for each candidate token of the generated hypotheses.
Hence, to compare the contextualized and decontextualized versions of the same system, we adjust the \seqObj{}'s alignment $\mathcal{A'}$ (Eq.\ \eqref{eq:contextual_align}) to utilize the \textit{decontextualized} embeddings (Eq.~\eqref{eq:dec_embeddings}) instead of the contextualized ones:
\begin{equation}
   \begin{aligned}
        \mathcal{L}_{\seqDecObj{}}(\Theta) &= \mathcal{L}_{\seqObj{}}(\Theta, \mathcal{A'}_{\mathit{dec}})
         \\
        \mathcal{A'}_{\mathit{dec}}(s_\Theta^i, t_2) &= \!\!\!\min_{e_{\mathit{dec}}^j \in  \Theta_{\mathit{dec}}(t_2)} \textsc{D}(e_{\mathit{dec}}^i, e_{\mathit{dec}}^j)
    \end{aligned}
    \label{eq:seq_align_dec}
\end{equation}
All other parameters of \seqObj{} remain unchanged, as described in Section~\ref{sec:experimental_setup}.

\paragraph{Impact of the loss formulation}
\label{sec:ablation:loss_formulation}

Following the previous work on sequential objectives (§\ref{sec:background}), \seqObj{} utilize the distance-based loss, but since we use token-level alignment, similarly to standard \MLObj{}, we could also formulate the objective using Cross Entropy (\textit{CE}).

This ablation evaluates the impact of the loss formulation by introducing an analogous objective to \seqDecObj{} (Eq.~\eqref{eq:seq_align_dec}), but utilizing the \textit{CE} loss instead of $L1$ distance:
\begin{equation}
    \mathcal{L}_\textit{SCE}(\Theta) \!= \min\!\left(\!\! - \log
    \!\frac{\!\exp(\Theta(X_j, \Pi_{1..i-1}^\Theta\!(X_j)))}
    {\exp(\mathcal{A}_\textit{dec}(\textit{voc}_\Theta, \!Y_j))}\!\right)
    \label{eq:sce}
\end{equation}
We sample the prefixes from the model's own hypotheses using the same generation strategy $\Pi^\Theta$ as in other sequential objectives.
We use the decontextualized objective as the reference to avoid the overhead of inference of contextual embeddings for the full vocabulary.

\section{Results}
\label{sec:results}

Table~\ref{table:results_overview} compares the results of adaptation using a selection of baseline methods and our two main objectives: \tokenObj{} and \seqObj{}, as trained on a selected domain and evaluated on a held-out set of the same domain (ID) and other domains (OOD). The domains are ordered by ascending size of the training data.
Table~\ref{table:ablation} additionally includes the objectives from our Ablation experiments. 
More detailed, per-domain ablations results can be found in Table~\ref{table:results_all} in Appendix~\ref{appx:detailed_results}.

\textbf{Alignment-based objectives improve robustness;}
Both \tokenObj{} and \seqObj{} consistently improve the model robustness (OOD) over the \MLObj{} in \textit{all} the evaluated cases and on average deliver more robust models compared to all other methods.
In addition, comparing \tokenObj{} to instances of \MLObj{}, we also see the advances in the adaptation quality (ID), in four out of five cases where \MLObj{} is able to deliver any ID improvements.
In OOD evaluations, \seqObj{} is slightly more robust than \tokenObj{}, presenting a more robust, yet also technically more complex alternative.


\begin{table}[tb]
\centering
\scalebox{1}{%
\begin{tabular}{@{}lll@{}}
\toprule
\multicolumn{1}{r}{$\Delta$BLEU:} & \multicolumn{1}{c}{ID}           & \multicolumn{1}{c@{}}{OOD}          \\
\midrule
0. \textit{\MLObj{}}             & \multicolumn{1}{l}{$+\,\ 8\%\pm31\%$} & \multicolumn{1}{l@{}}{$-27\%\pm29\%$} \\
1. \tokenObj{}        & \multicolumn{1}{l}{$+\ \,8\%\pm30\%$}            & \multicolumn{1}{l@{}}{$-\ \,3\%\pm\ \,9\%$}                                 \\
2. \seqObj{} & \multicolumn{1}{l}{$+\,\ 3\%\pm27\%$}                     & \multicolumn{1}{l@{}}{$-\ \,1\%\pm\ \,8\%$}                                 \\
\hline 
3. \seqRandomObj{}        &  \multicolumn{1}{l}{$+\,\ 3\%\pm31\%$}        & \multicolumn{1}{l@{}}{$-\,\ 6\%\pm\ \,5\%$}                                 \\
4. \seqDecObj{}  & \multicolumn{1}{l}{$+\,\ 5\%\pm31\%$} & \multicolumn{1}{l@{}}{$-\,\ 6\%\pm27\%$} \\

5. \seqCEObj{}  &  \multicolumn{1}{l}{$+\,\ 4\%\pm32\%$}                & \multicolumn{1}{l@{}}{$-17\%\pm44\%$} \\
\bottomrule
\end{tabular}}
\caption{\textbf{Results of Ablation experiments}: Average change of BLEU scores relative to the original model, when adapting the Transformer-base model with a given objective.
The intervals cover the averages of 6~in-domain and 20~out-of-domain evaluations (§\ref{sec:experimental_setup}).
\vspace{-2mm}
}
\label{table:ablation}
\end{table}


While the average results confirm our main hypothesis that circumventing \MLObj{}'s assumption of a single-truth prediction can improve the model's distributional robustness, we see a large variance in the performance of our methods similar to \MLObj{}. 
The in-domain results of \seqObj{} also dispute our assumption that self-generation of prefixes could compensate for the scarcity of natural in-domain data; \seqObj{}'s ID performance on the two smallest domains is inferior to both \MLObj{} instances, while it is very efficient in the higher-resource \textit{Law/DGT}.

\textbf{Avoiding teacher-forcing improves robustness;}
A~comparison of the results of \seqRandomObj{} and \MLObj{} in Table~\ref{table:ablation} shows that the mere exposition of the model to its own hypotheses reduces the forgetting of \MLObj{} by 77\% in average ($-27\%\rightarrow -6\%$). However, constructing the non-informative targets for self-generated inputs also causes a decay in adaptation quality ($+8\%\rightarrow +3\%$).

\textbf{Alignment-based targets complement avoiding teacher-forcing;}
Robustness improvements of \seqObj{} over \seqRandomObj{} (Table~\ref{table:ablation}) might be attributed to the semantically-grounded Alignment targets (§\ref{sec:alignment}).
While the aggregate in-domain results of \seqObj{} and \seqRandomObj{} in Table~\ref{table:ablation} are very close, the per-domain results (Table~\ref{table:results_all} in Appendix~\ref{appx:detailed_results}) reveal that their results vary over domains and the suggested ID tie of \seqRandomObj{} to \seqObj{} is largely attributed to \seqRandomObj{}'s better results on \textit{Bible}, where both objectives fail to improve ID nevertheless.
\looseness=-1

\textbf{Decontextualization does not carry a large qualitative drop;}
Both objectives grounding their targets in decontextualized embeddings (\tokenObj{} and \seqDecObj{}) show relatively good average results on both ID and OOD (Table~\ref{table:ablation}), but \tokenObj{} is the only method reaching adaptation accuracy comparable to \MLObj{} in average.
A comparison of \seqObj{} to its decontextualized instance (\seqDecObj{}) specifically evaluates the impact of decontextualization, in the settings of absolute distance loss and no teacher forcing.
We see that while the decontextualization leads to a larger loss in the robustness ($-1\%\rightarrow -6\%$), \seqDecObj{} slightly outperforms \seqObj{} on the in-domain ($+3\%\rightarrow +5\%$). 
Per-domain results (Table~\ref{table:results_all} in Appendix~\ref{appx:detailed_results}) show that this is attributed mainly to the superior adaptation performance of \seqDecObj{} in the low-resource \textit{Opensubs (en→ukr)} case, suggesting that the embeddings' averaging within decontextualization (§\ref{sec:decontextualization}) works well also with small amounts of texts. 

\textbf{Loss formulation impacts model robustness;}
A~comparison of \seqDecObj{} and \seqCEObj{} in Table~\ref{table:ablation} assesses the impact of changing objectives' loss formulation from $L1$ to Cross Entropy (CE). We see that changing a distance-based loss to CE causes a significant drop in OOD robustness ($-6\%\rightarrow-17\%$), comparable to the drop of the traditional \MLObj{}, also built upon CE loss ($-21\%$). However, the superior OOD performance of CE-based \tokenObj{} contradicts that CE loss itself could be a~pivotal cause of catastrophic forgetting.

\section{Conclusion}
\label{sec:conclusion}

Our work sets out to explore the alternatives between the efficient yet na\"{\i}ve \MLObj{} objective and expressive but resource-demanding sequential objectives, by building the training signal from the semantic token representations.
We build an alignment mechanism applicable with an arbitrary representation model and propose objectives that utilize a domain-agnostic embedding model as its target. We find that using semantically-grounded targets in adaptation persists robustness of the model much better than other methods, without compromises in in-domain performance.

We additionally explore the impact of selected design choices on the robustness of generative LLMs in the ablation experiments. Among others, we find that a major part of the model's robustness can be persisted merely by including the model's own outputs among the inputs, attributing a part of adaptation forgetting to exposure bias.
Future work might also build upon the qualitative assessment of the impact of decontextualization, resolving the computational overhead of applying the contextualized embeddings in dynamic contexts.

We look forward to future work that will explore the potential of applying semantically-grounded objectives in a more robust and data-efficient training of LLMs for many other applications, including the pre-training stages. 

While our experiments do not evaluate such settings, we note that our methods complement the model-centric ones, including recent parameter-efficient training strategies \citep{valipour-etal-2023-dylora,dettmers2023qlora}. Given the encouraging results of \textit{LoRA} (Table~\ref{table:results_overview}), we believe that future work combining parameter-efficient methods with semantically-grounded objectives like ours can mitigate forgetting of domain and task adaptation even further.

\section*{Limitations}
\label{sec:limitations}

We experiment with a range of adaptation domains that we draw systematically to capture the covariates enumerated in Section~\ref{sec:datasets}.
However, future work should acknowledge that these are not all the covariates responsible for the success of adaptation and the robustness of the final model. Following is the non-exhaustive list of possible covariates that we do not control in this work.
(i)~the adapted model size,
(ii)~the size of pre-training data,
(iii)~pre-training configuration parameters, but also
(iv)~the broad variance of adapted language pair(s);
(v)~the variance of mutual similarity of languages within the pair, and hence 
(vi)~the difficulty of training the translation model.

The evaluation of our experiments did not consider the effect of \textit{randomness} of the training process. Despite the fact that our experiments were run with a fixed random seed and initial value, making our results deterministically reproducible, the variance of the results among the experiments of different random seeds was not investigated due to the related infrastructural costs. However, all our results are aggregated over a larger set of checkpoints and/or domains, ranging from 10 (IDs in Table~\ref{table:results_overview}) to 720 (OODs in Table~\ref{table:ablation}), as described in Section~\ref{sec:experimental_setup}.

The alignment scheme proposed in Section~\ref{sec:alignment} might have blind spots; for instance, in the cases utilizing decontextualized embeddings, where both the hypothesis and reference contain multiple occurrences of the same word, the alignment scheme will make the prediction of the same target token equally \textit{good}, regardless of the position.
In future work, this imperfection could be addressed by using the Optimal transport algorithm \citep{kusner2015from} within the Alignment, similarly to \citet{Zhang2020SemanticMF}.

\subsection*{Acknowledgements}
We thank the anonymous reviewers of our work for providing us with qualified feedback that significantly shaped the resulting form of this paper.

We acknowledge the Centre for Biomedical Image Analysis at Masaryk University supported by MEYS CR (LM2018129 and CZ.02.1.01/0.0/0.0/18\_046/0016045 Czech-BioImaging) for their support in obtaining the results presented in this paper.

\bibliography{stefanik}
\bibliographystyle{acl_natbib}

\appendix

\section{Hyperparameters}
\label{appx:hyperparameters}

For each of the evaluated objectives, we perform a hyperparameter search independently over the selected parameters in the denoted range, based on the best in-domain validation BLEU reached in the adaptation to \textit{Wikimedia} domain.

(1)~\textbf{learning rate}: ranging from $2\cdot 10^{-7}$ to $2\cdot10^{-4}$, with step~10. (2)~\textbf{objectives ratio $\alpha$} (Eq.\ \eqref{eq:weighting}): we manually set the weight of the additional objective such that the loss values for both components of the final loss are approximately balanced, based the first 10~valuations.
We do not perform further tuning and use the same weights over all experiments.
(3)~\textbf{Batch size}: For \textit{ML} experiments, we fix the effective batch size to $60$, we pick the optimal batch size for \tokenObj{} and \seqObj{} objectives over $[1, 5, 10, 20]$.

Other parameters that we adjust and remain fixed over the experiments are the following: $\textbf{warmup\ steps}=1,000$, \textbf{LR schedule} as \textit{constant decay}. Distance-based objectives including \seqObj{} introduce two new parameters: 
(i)~$K$: a number of the sampled hypotheses and 
(ii) $n$: a number of most-likely tokens to align.
To keep the computation time feasible, we do not perform further tuning and set these parameters to $K=10$ and $n=3$ over all the experiments. 
All other parameters can be retrieved from the defaults of TrainingArguments of Transformers \cite{Wolf2019HuggingFacesTS}, version 4.10.2.

We treat the optimized hyperparameters as \textit{independent}; hence we optimize each variable separately.
Our configuration results in experimenting with 9~hyperparameter search runs for each objective, including \MLObj{} baseline.

We also tune selected parameters of \textit{Adapters} and \textit{LoRA} implementations based on their original papers: (i)~A~compressed representation size ratio $\frac{t}{h}$ to model hidden state size~$h$ is chosen from $t \in [2, 4, 16, 32]$, (ii)~a~learning rate is chosen from $\text{LR} \in [2\cdot10^{-3}, 2\cdot10^{-4}, 2\cdot10^{-5}]$. We pick as optimal $h=32$, $h=16$ and 
$\text{LR}=2\cdot10^{-4}$, $\text{LR}=2\cdot10^{-5}$ for \textit{Adapters} and \textit{LoRA}, respectively.

\section{Computational Requirements}
\label{appx:speed}

We performed the adaptation of each of the proposed objectives on a server with a single \textit{NVidia Tesla A100}, 80\,GB of graphic memory, 512\,GB of RAM and 64-core processor (\textit{AMD EPYC 7702P}).
We also tested to train all our experiments using lower configuration using a single \textit{NVidia Tesla T4}, 16\,GB of graphic memory, 20\,GB of RAM, and a single core of \textit{Intel(R) Xeon(R)} processor.

We benchmark the running times of the time-demanding parts of the adaptation process in the first-mentioned configuration.
We find that the proposed decontextualization process required by \tokenObj{}, \seqCEObj{} and \seqDecObj{} takes in these settings between 50~minutes on the smallest domain to 25~hours on the largest domain. 
Table~\ref{table:speed} shows the average speed of updates and the number of steps that each of the designed objectives requires to converge.
Further details on our methodology are described in Section~\ref{sec:experimental_setup}.

\begin{table}[tb]
\centering
\scalebox{0.96}{%
\begin{tabular}{@{}l@{\,}cc@{}}
\toprule
Objective    & Updates / hour & Updates to converge \\
\midrule
\MLObj{}     & 451            &  15,500    \\
\midrule
\tokenObj    & 404            &  24,000    \\
\seqObj{}    & 287            &  11,875    \\
\seqRandomObj& 152            &  10,100    \\
\seqDecObj{} & 295            &  \hphantom{0}7,500\\
\seqCEObj{}  & 585            &  23,740 \\
\bottomrule
\end{tabular}
}
\caption{\textbf{Adaptation speed}: Average number of updates per hour and average number of updates to converge that we measure over objectives in our experiments.}
\label{table:speed}
\end{table}

\section{Details of Alignment Algorithm}
\label{appx:alignment}

Algorithm \ref{algo:align_to_grounding} describes the alignment procedure that we propose to obtain \textit{grounding embeddings} for the tokens of the trained model.

Our approach first \textit{aligns} the model and embeddings vocabulary; Given a text~$t$, we obtain two ordered sequences of textual subwords (tokens): grounding embeddings tokens $s_e(t)$ and model tokens $s_\Theta(t)$. We obtain the \textit{model grounding embeddings} $e_\Theta^i$ of each \textit{model} subword $s_\Theta^i \in s_\Theta(t)$ to each \textit{grounding} subword $s_{e,i} \in s_\Theta(t)$ by 
(i)~assigning the \textit{coverage intervals} of $t$ to each model and embedding subword $s_\Theta(t)$ and $s_e(t)$, and
(ii)~for each model subword \mbox{$s_\Theta^i \in s_\Theta(t)$}, searching for the subword $s^i_e(t)$ with \textit{largest intersection} of the covering intervals $|s_\Theta^i \cap s_e^j|$.

\begin{algorithm}[tbh]
\SetKwProg{proc}{proc}{}{}
\let\var\relax
\scalebox{1.1}{%
\begin{minipage}{1.06\linewidth}
\setstretch{1.05}
\proc{\upshape $\textit{align\_to\_grounding}(\var{s_\Theta}, \var{s_e})$:}{%
    \ForEach{$i \in 1..|\var{s_\Theta}|$}{
        \While{$|\var{s_\Theta^i} \cap \var{s_e^j}| > \var{best\_cov}\ $}{
            $\var{pair\!s_i} \gets j$\\
            $\var{best\_cov} \gets |\var{s_\Theta^i} \cap \var{s_e^j}|$ \\
            $j \gets j+1$
        }
    }
    \Return{$\var{pair\!s}$}
}
\vspace*{1mm}
\end{minipage}
}
\caption{Ability to pair each model token $s_\Theta^i$ with the best-matching grounding subword $s_e^j$ allows us to use alignment grounded in domain-agnostic representations.
Relying on the consistent ranking of the aligned sequences, the grounding alignment algorithm requires at most $(|s_{\Theta}| + |s_{e}|)$ steps to finish.}
\label{algo:align_to_grounding}
\end{algorithm}

\begin{figure}[t]
\centerline{\includegraphics[width=1.08\linewidth]{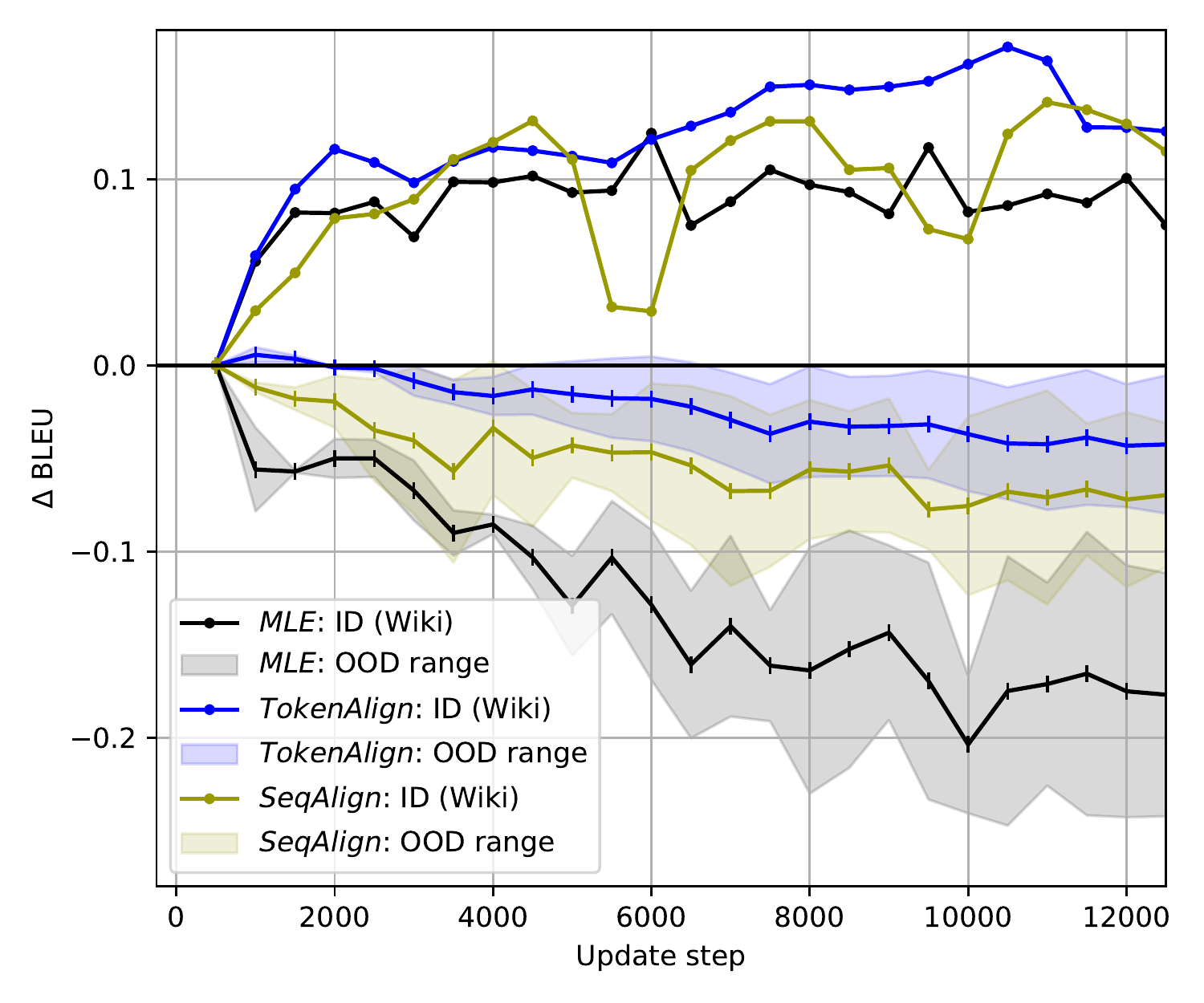}}%
\vspace*{-.5\baselineskip}
\caption{In-domain (ID) and out-of-domain (OOD) change of the original BLEU in domain adaptation of a translation model using \textbf{\MLObj{}} and the two introduced objectives: \color{Lightblue}\textbf{\tokenObj{}} \color{black}and \color{Yellow}\textbf{\seqObj{}}\color{black}. Adaptation of Transformer-base model on Wikipedia, evaluated on a held-out set of the adapted domain (in-domain, ID) and a variety of out-of-domain (OOD) datasets (§\ref{sec:experimental_setup}).}
\label{fig:abstract_old}
\end{figure}

\section{Detailed Results of Ablation Objectives}
\label{appx:detailed_results}

\addtocounter{figure}{1}
\begin{table*}[p]
\centering
\scalebox{0.9}{%
\begin{tabular}{@{}rrlllll@{}}
\toprule
& \multicolumn{1}{l}{$\Delta$ BLEU} &
  \multicolumn{1}{l}{\begin{tabular}[c]{@{}l@{}} Bible\\ (de→en)\end{tabular}} &
  \multicolumn{1}{l}{\begin{tabular}[c]{@{}l@{}} Opensubs\\ (en→ukr)\end{tabular}} &
  \multicolumn{1}{l}{\begin{tabular}[c]{@{}l@{}} Wiki\\ (en→cze)\end{tabular}} &
  \multicolumn{1}{l}{\begin{tabular}[c]{@{}l@{}} Medical/EMEA\\ (est→en)\end{tabular}} &
  \multicolumn{1}{l}{\begin{tabular}[c]{@{}l@{}} Law/DGT\\ (en→de)\end{tabular}} \\
    & & \small{50,000 pairs} & \small{80,000 pairs} & \small{100,000 pairs} & \small{300,000 pairs} & \small{5,100,000 pairs} \\

  \hline
 Orig. BLEU & & $21.89$ & $26.12$ & $34.04$ & $54.85$ & $33.56$ \\
\midrule
\MLObj{}         & ID    & $-\,\ 8\%$                   & $+\,\ 4\%$                  & $\,\ +9\%$                    & $+38\%$                     & $-\,\ 1\%$                      \\
                 & OOD   & $-53\%\pm36\%$            & $-15\%\pm9\%$           & $-15\%\pm5\%$             & $-35\%\pm10\%$              & $-19\%\pm11\%$               \\ \hline
\tokenObj{}      & ID    & $-21\%$                   & $+\,\ 8\%$                  & $\boldsymbol{+12\%}$           & $\boldsymbol{+45\%}$            & $+\,\ 1\%$                      \\
                 & OOD   & $-\,\ 2\%\pm1\%$            & $\boldsymbol{-\,\ 1\%\pm1\%}$  & $\boldsymbol{-\,\ 6\%\pm6\%}$    & $-\,\ 6\%\pm7\%$              & $\boldsymbol{+\,\ 6\%\pm20\%}$      \\ \hline
\seqObj{}        & ID    & $-23\%$                   & $-\,\ 8\%$                  & $+\,\ 8\%$                    & $+31\%$                     & $\boldsymbol{+\,\ 7\%}$             \\
                 & OOD   & $\boldsymbol{-\,\ 1\%\pm1\%}$   & $-\,\ 2\%\pm3\%$           & $-12\%\pm5\%$             & $\boldsymbol{-\,\ 1\%\pm2\%}$     & $+\,\ 3\%\pm13\%$               \\ \hline
\seqRandomObj{}  & ID    & $-14\%$                   & $-\,\ 7\%$                  & $+\,\ 8\%$                    & $+34\%$                     & $-\,\ 7\%$                      \\
                 & OOD   & $-\,\ 8\%\pm2\%$            & $-\,\ 3\%\pm3\%$           & $-\,\ 9\%\pm3\%$             & $-\,\ 7\%\pm5\%$              & $-\,\ 7\%\pm5\%$               \\ \hline
\seqDecObj{}     & ID    & $-26\%$                   & $\boldsymbol{+11\%}$         & $+\,\ 5\%$                    & $+35\%$                     & $+\,\ 2\%$                      \\
                 & OOD   & $-13\%\pm8\%$            & $-\,\ 1\%\pm1\%$           & $-11\%\pm19\%$               & $-12\%\pm7\%$              & $+\,\ 4\%\pm17\%$               \\ \hline
\seqCEObj{}      & ID    & $\boldsymbol{+\,\ 8\%}$          & $+\,\ 9\%$                  & $+11\%$                    & $+\,\ 1\%$                     & $-11\%$                      \\
                 & OOD   & $-78\%\pm9\%$            & $-32\%\pm1\%$           & $-12\%\pm5\%$             & $-\,\ 1\%\pm2\%$              & $-14\%\pm13\%$               \\
\bottomrule

\end{tabular}
}
\caption{\textbf{Evaluation of adaptation quality and robustness over \textit{all} designed objectives}:
A~change of BLEU score relative to the original model, when adapting pre-trained Transformer-base on a selected domain, as measured on a test set of the training domain (in-domain, ID) and out-of-domain (OOD).
The aggregates over all domains are listed in Table~\ref{table:ablation}.}
\label{table:results_all}
\bigskip

\centering
\scalebox{0.9}{
\includegraphics[width=\textwidth]{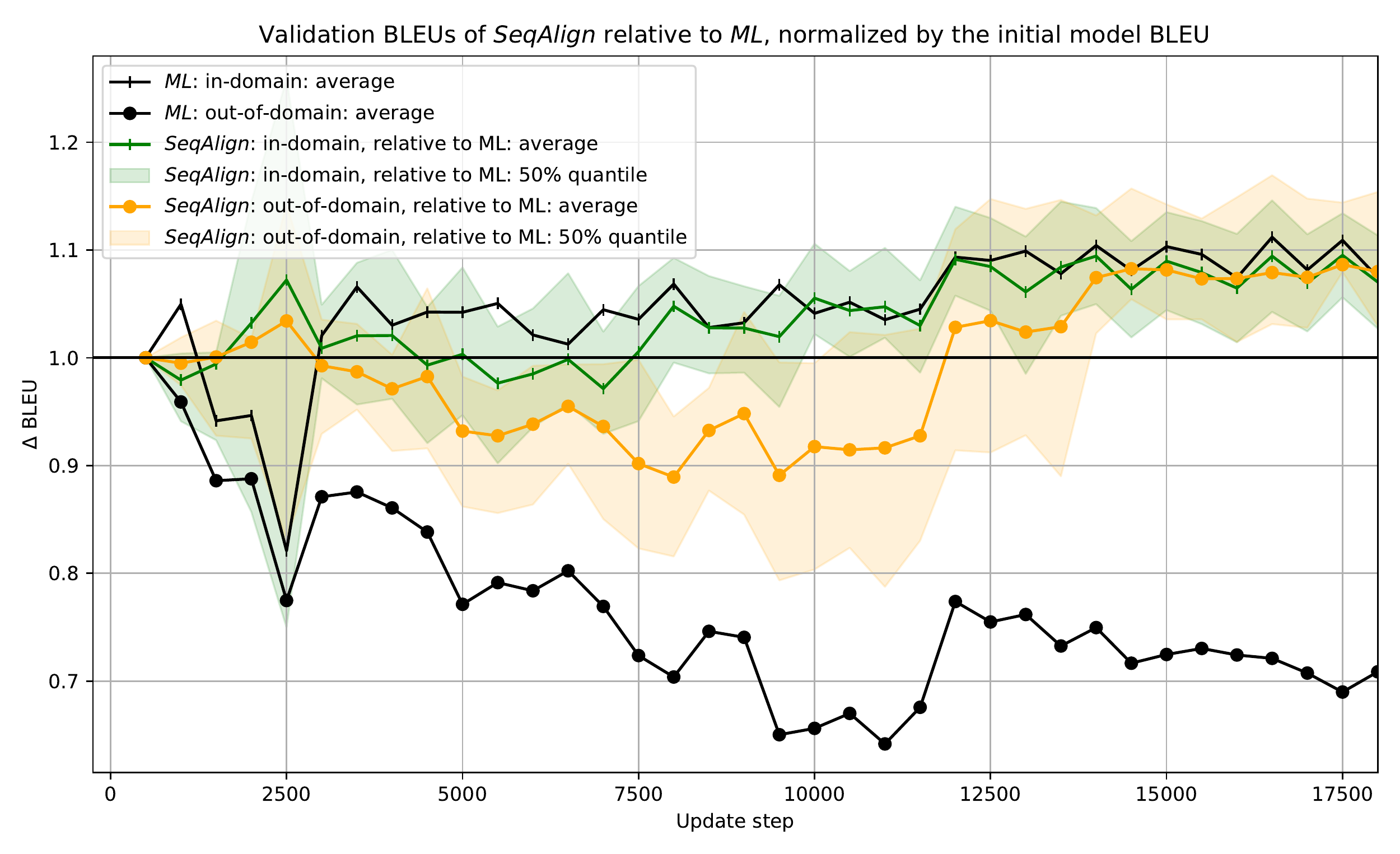}
}
\vspace*{-.5\baselineskip}
\let\tablename\figurename
\addtocounter{table}{1}
\caption{Comparison of \textbf{validation BLEU of \MLObj{} and \seqObj{} objective} reported over the training on 5~different domains and 20~corresponding out-of-distribution domains until the in-domain early-stopping. For easier comparison, both \MLObj{} logs are averaged, and reported intervals correspond to the 50\%-quantile of difference to the \MLObj{} run on the corresponding evaluation domain. While the training with \MLObj{} objective consistently magnifies the \textit{forgetting} of adaptation, the soft objectives report a higher OOD score over all experiments while reaching comparable adaptation gains on the in-domain. 
Note that the two major gains of \seqObj{} before steps 12,000 and 14,000 are attributed to the early stopping of specific runs at these points and hence, should be excluded from the conclusions. See Appendix~\ref{appx:training_logs} for further description.}
\label{fig:training_loss_seq}
\end{table*}


Table \ref{table:results_all} shows a comparison of \textit{all} objectives over all evaluated domains, providing a finer-grained report of results presented in Table~\ref{table:ablation}.
Note that in order to eliminate the effect of different scaling of BLEU evaluations in character-segmented BLEU results, we exclude the (en→zh) pair from the ablations.
The methodology of results collections is described in Section~\ref{sec:experimental_setup}.
The discussion including these results is present in Section~\ref{sec:results}.

\section{Training Validation Reports}
\label{appx:training_logs}

\begin{figure*}[t]
\centering
\scalebox{0.9}{%
\includegraphics[width=\textwidth]{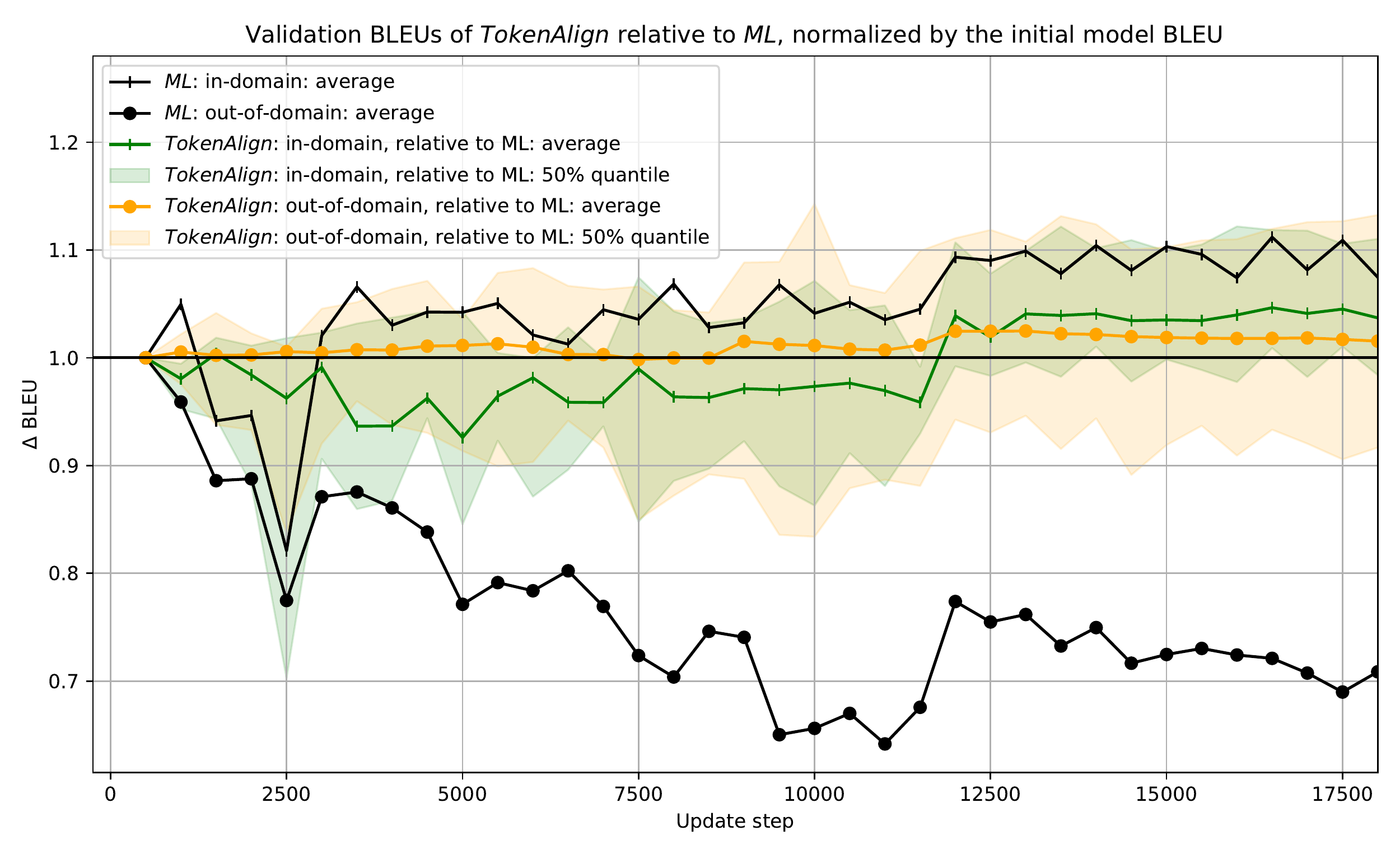}
}
\vspace*{-.5\baselineskip}
\caption{Comparison of \textbf{validation BLEU of \MLObj{} and \tokenObj{} objective} as reported over the training on 5~different domains and 20~corresponding out-of-distribution domains until in-domain early-stopping. See Figure~\ref{fig:training_loss_seq} and Appendix~\ref{appx:training_logs} for further description.}
\label{fig:training_loss_tbert}
\end{figure*}

We report and compare the change of validation BLEU of our two main objectives, relative to the \MLObj{} objective over the course of our experiments and overview the results in Figures~\ref{fig:training_loss_seq} and \ref{fig:training_loss_tbert} for \seqObj{} and \tokenObj{} objective, respectively. A~comparison of all three objectives is in Figure~\ref{fig:abstract_old}.
\onecolumn\begin{multicols}{2} 

The plots aggregate 5~training logs and their corresponding out-of-domain logs into the in-domain and out-of-domain reports, for easy comparability with \MLObj{}, both in-domain and out-of-domain BLEUs of \MLObj{} are \textit{averaged} and paired with the corresponding BLEUs of the inspected objective over the shared evaluation domain. Finally, the plots of the inspected objective consist of \textit{50\% quantile intervals} and the \textit{average} of BLEU relative to both the \MLObj{} BLEU and initial model performance.
Note that while the relative distances of \MLObj{} to the corresponding plots of the other objective \textit{always} correspond, some training runs are terminated in the course of the plotted steps, explaining some sudden performance gains in the plot.

While the performance decay of \MLObj{} by the time of early-stopping by in-domain BLEU is close to linear, \tokenObj{} on average maintains none, or minimal decays of the out-of-domain performance, although the variance of the initial decay significantly varies over domains. This trend implies that the early-stopping strategy based on in-domain performance does not significantly decay the robustness results and favors the deployment of \tokenObj{} in situations where no validation out-of-domain data is present.

The robustness of the model trained using \seqObj{} behaves differently and the initial robustness decay is more significant. However, the decay soon diverges from \MLObj{} and noticeably, after the \mbox{5,000-th} step \textit{all} the robustness evaluations of \seqObj{} report robustness gains over \MLObj{}.

Although we restrain from drawing conclusions based exclusively on these plots, the comparisons suggest that while the decay of robustness of \MLObj{} training is continuous, in the case of soft objectives, the decay gradually slows, while the model incrementally reaches potential in-domain gains similar to \MLObj{}.
\end{multicols}

\end{document}